%% file: paper-arxiv.tex
\documentclass{article}
\PassOptionsToPackage{numbers, compress}{natbib}
\usepackage[final]{nips_2016}

\usepackage[utf8]{inputenc} % allow utf-8 input
\usepackage[T1]{fontenc}    % use 8-bit T1 fonts
\usepackage{hyperref}       % hyperlinks
\usepackage{url}            % simple URL typesetting
\usepackage{booktabs}       % professional-quality tables
\usepackage{amsfonts}       % blackboard math symbols
\usepackage{nicefrac}       % compact symbols for 1/2, etc.
\usepackage{microtype}      % microtypography

\input{preamble}
\input{preamble-notation-new}

\title{Similarity-based Multi-label Learning}

\author{
Ryan A. Rossi\\
Palo Alto Research Center\\
\texttt{rrossi@parc.com}\\
\And
Nesreen K. Ahmed\\
Intel Labs\\
\texttt{nesreen.k.ahmed@intel.com}\\
\AND
Hoda Eldardiry\\
Palo Alto Research Center\\
\texttt{heldardiry@parc.com}\\
\And
Rong Zhou\\
Google\\
\texttt{rongzhou@google.com}\\
}

\begin{document}

\maketitle

\begin{abstract}
%\vspace{-3mm}
Multi-label classification is an important learning problem with many applications.
In this work, we propose a principled similarity-based approach for multi-label learning called $\sml$. 
We also introduce a similarity-based approach for predicting the label set size.
The experimental results demonstrate the effectiveness of $\sml$ for multi-label classification where it is shown to compare favorably with a wide variety of existing algorithms across a range of evaluation criterion.
\end{abstract}

\section{Introduction}\label{sec:intro}
Multi-label classification is an important learning problem~\cite{tsoumakas2006multi} with applications in 
bioinformatics~\cite{pavlidis2000combining},
image \& video annotation~\cite{carneiro2007supervised,wang2009multi} and
query suggestions~\cite{agrawal2013multi}.
The goal of multi-label classification is to predict a label vector $\vy \in \{0,1\}^{\k}$ for a given unseen data point $\vx \in \RR^{\m}$.

Previous work has mainly focused on reducing the multi-label problem to a more standard one such as 
multi-class~\cite{mccallum1999multi,boutell2004learning} and binary classification~\cite{boostexter}, ranking~\cite{ranksvm} and regression~\cite{ji2009multi,hsu2009multi};~see~\cite{zhang2014review} for a recent survey.
Standard multi-class approaches can be used by mapping a multi-label problem with $K$ labels to a multi-class problem with $2^{\k}$ labels~\cite{mccallum1999multi,boutell2004learning}.
Binary classification methods can also be used by copying each feature vector $\k$ times and for each copy $k$ an additional dimension is added with value $k$; and the training label is set to $1$ if label $k$ is present and 0 otherwise~\cite{boostexter}.
Rank-based approaches attempt to rank the relevant labels higher than irreverent ones~\cite{ranksvm}.
Regression methods map the label space onto a vector space where standard regression methods can be applied~\cite{ji2009multi,hsu2009multi}.

In this work, we introduce a similarity-based approach for multi-label learning called $\sml$ that gives rise to a new class of methods for multi-label classification.
Furthermore, we also present a similarity-based set size prediction algorithm for predicting the number of labels associated with an unknown test instance $\vx$.
Experiments on a number of data sets demonstrate the effectiveness of $\sml$ as it compares favorably to existing methods across a wide range of evaluation criterion.
The experimental results indicate the practical significance of $\sml$.

In addition, $\sml$ is a direct approach for multi-label learning.
This is in contrast to existing methods that are mostly \emph{indirect approaches}
that transform the multi-label problem to a binary, multi-class, or regression problem and apply standard algorithms (\eg, decision trees).
Furthermore, other rank-based approaches such as \textsc{Rank-svm}~\cite{ranksvm} 
are also indirect extensions 
of \textsc{svm}~\cite{vladimir1995nature,wolfe1961duality} to multi-label classification.
Notably, $\sml$ completely avoids such mappings (required by \textsc{svm}) and is based on the more general notion of similarity.

\section{Preliminaries} 
\label{sec:prelim}
Let $\mathcal{X}=\RR^{\m}$ denote the input space and let $\mathcal{Y} = \{1,2,\ldots,\k\}$ denote the set of possible class labels.
Given a multi-label training set $\mathcal{D}$ defined as:
$\mathcal{D} = \{(\vx_1, Y_1), \ldots, (\vx_\n, Y_\n)\}$
\noindent
where $\vx_i \in \mathcal{X}$ is a $\m$-dimensional training vector representing a single instance and $Y_i$ is the label set associated with $\vx_i$.
Given $\mathcal{D}$ the goal of the multi-label learning problem is to learn a function $h : \mathcal{X} \rightarrow 2^{\k}$ which predicts a set of labels for an unseen instance $\vx_j \in \RR^{\m}$.
A multi-label learning algorithm typically outputs a real-valued function $f : \X \times \Y \rightarrow \RR$ where $f_k(\vx_i)$ is the confidence of label $k \in \Y$ for the unseen test instance $\vx_i$.
Given an instance $\vx_i$ and its associated label set $Y_i$, a good multi-label learning algorithm will output larger values for labels in $Y_i$ and smaller values for labels not in $Y_i$.

We consider a variety of evaluation criterion for comparing multi-label learning algorithms.
The multi-label hamming loss is the fraction of incorrectly classified instance-label pairs:
\begin{equation} \label{eq:eval-hamming-loss}
\mathbb{E}_{\D}(f) = \frac{1}{\n} \sum_{i=1}^{\n} \; \frac{1}{\k} \Big|\, h(\vx_i)\, \Delta\, Y_i \Big|
\end{equation}
\noindent
where $\Delta$ is the symmetric difference between the predicted label set $\hat{Y}_i = h(\vx_i)$ and the actual ground truth label set $Y_i$.
One-error evaluates how many times the top-ranked label is not in the set of ground truth (held-out) labels:
\begin{equation}\label{eq:eval-one-error}
\mathbb{E}_{\D}(f) = \frac{1}{\n} \sum_{i=1}^{\n} \; 
\mathbb{I}\Bigg[\, \big[\argmax_{k \in \Y} f_k(\vx_i)\big] \not\in Y_i \,\Bigg]
\end{equation}
\noindent
where for any predicate $p$ the indicator function $\mathbb{I}[\,p\,] = 1$ iff $p$ holds and $0$ otherwise.
Given a set of labels ordered from most likely to least, coverage measures the max position in the ordered list such that all proper labels are recovered:
\begin{equation} \label{eq:eval-coverage}
\mathbb{E}_{\D}(f) = \frac{1}{\n} \sum_{i=1}^{\n} \; \max\limits_{k \in \Y} \; \pi(\vx_i, k) - 1
\end{equation}
\noindent
where $\pi(\vx_i, k)$ is the rank of label $k \in \Y$.
Alternatively, \emph{Ranking loss} measures the fraction of reversely ordered label pairs:
\begin{equation}\label{eq:eval-ranking-loss}
\! \mathbb{E}_{\D}(f) \! = \! \frac{1}{\n} \! \sum_{i=1}^{\n} \frac{1}{|Y_i||\bar{Y}_i|}  \Big|\!
\Big\lbrace \! (k, k^{\prime}) \! \in \! Y_i \! \times \! \bar{Y}_i \; \big| \; f_k(\vx_i) \! \leq \! f_{k^{\prime}}(\vx_i) \!\Big\rbrace
\!\Big|
\end{equation}
\noindent%
Average precision measures the average fraction of relevant labels ranked higher than a particular label $k \in Y_i$:
\begin{equation}\label{eq:eval-avg-precision}
\mathbb{E}_{\D}(f) = \frac{1}{\n} \sum_{i=1}^{\n} \frac{1}{|Y_i|}
\sum_{k \in Y_i} 
\frac{\big| \big\{ k^{\prime} \! \in \! Y_i \; | \; \pi(\vx_i, k^{\prime}) \! \leq \! \pi(\vx_i, k) \!\big\}\big|}{\pi(\vx_i, k)}
\end{equation}
\noindent%
Multi-label algorithms should have high precision (Eq.~\ref{eq:eval-avg-precision}) 
with low hamming loss (Eq.~\ref{eq:eval-hamming-loss}), 
one-error (Eq.~\ref{eq:eval-one-error}), 
coverage (Eq.~\ref{eq:eval-coverage}), and 
ranking loss (Eq.~\ref{eq:eval-ranking-loss}).

%======================================
% SML APPROACH
%======================================
\section{Similarity-based Multi-label Learning ($\sml$)}
This section presents our general similarity-based approach for multi-label learning called $\sml$.
Given a multi-label training set 
$\mathcal{D} = \{(\vx_1, Y_1),\ldots, (\vx_j, Y_j), \ldots, (\vx_\n, Y_\n)\}$
where $\vx_j \in \RR^{\m}$ is a $\m$-dimensional training vector representing a single instance and $Y_j$ is the label set associated with $\vx_j$, 
the goal of multi-label classification is to predict the label set $Y_i$ 
of an unseen instance $\vx_i \in \RR^{\m}$.
Assume $w.l.o.g.$ that all feature vectors $\vx_1,\ldots, \vx_\n$ are normalized to length 1.
Given the subset $\D_k \subseteq \D$ of training instances with label $k \in \{1,2,\ldots,\k\}$ defined as
\begin{equation} \label{eq:training-instances-with-label-k}
\D_k = \big\{(\vx_i,Y_i) \in \D \; | \; k \in Y_i\big\} 
\end{equation}
\noindent
we estimate the weight $f_k(\vx_i)$ of label $k$ for an unseen test instance $\vx_i \in \RR^{\m}$ as:
\begin{align} \label{eq:est-weight-for-label-k}
f_k(\vx_i) \; = \sum_{\vx_j \in \mathcal{D}_k} \simf\inner{\vx_i}{\vx_j}
\end{align}\noindent
where $\phi$ is an arbitrary similarity function.
Notably, the proposed family of similarity-based multi-label learning algorithms can leverage any arbitrary similarity function $\phi$. 
Furthermore, our approach does not require mappings in high-dimensional Hilbert spaces~\cite{weston1999support,hsu2002comparison} as required by \textsc{Rank-svm}~\cite{ranksvm}.
We define a few parameterized similarity functions below.
Given $\m$-dimensional vectors $\vx_i$ and $\vx_j$, the RBF similarity function is:
\begin{equation}\label{eq:rbf}
\simf(\vx_i, \vx_j) = \mathrm{exp}\Big[-\!\gamma \; \|\vx_i- \vx_j\|^{2} \, \Big]
\end{equation}
A common class of similarity measures for vectors of uniform length are polynomial functions: 
\begin{equation}
\label{eq:poly-kernel-nonlinear}
\simf(\vx_i,\vx_j) = \Big[\!\inner{\vx_i}{\vx_j} + c \Big]^{d}
\end{equation}
\noindent 
where $\inner{\cdot}{\cdot}$ is the inner product of two vectors, $d$ is the degree of the polynomial, and $c$ is a regularization term trading off higher-order terms for lower-order ones in the polynomial.
Linear-$\sml$ and quadratic-$\sml$ are special cases of Eq.~\eqref{eq:poly-kernel-nonlinear} 
where $d=1$ and $d=2$, respectively.
Furthermore, all label weights denoted by $f(\vx_i)$ for test instance $\vx_i$ are estimated as:
\begin{align} \label{eq:est-weight-for-all-labels}
\displaystyle
&f(\vx_i) \;=\;
\left[ \begin{array}{c}
\vphantom{\sum\limits_{\vx_j \in \D_1} \simf(\vx_i,\vx_j)} \\[-1.6em]
f_1(\vx_i) \\[0.8em] 
\vdots \\[0.6em]
f_{\k}(\vx_i) \\[0.8em]
\end{array} \right] 
\;=\;
\left[ \begin{array}{c}
\vphantom{\sum\limits_{\vx_j \in \D_1} \simf(\vx_i,\vx_j)} \\[-1.6em]
\sum\limits_{\vx_j \in \D_1} \simf\inner{\vx_i}{\vx_j} \\[0.8em]
\vdots \\[0.6em]
\sum\limits_{\vx_j \in \D_{\k}} \simf\inner{\vx_i}{\vx_j}\\[0.8em]
\end{array} \right] 
\end{align}\noindent

After estimating $f(\vx_i) = \big[ \, f_1(\vx_i) \; \cdots \; f_\k(\vx_i) \,\big]^{T} \in \RR^{\k}$ via Eq.~\ref{eq:est-weight-for-all-labels}, we predict the label set $Y_i$ of $\vx_i$; see Section~\ref{sec:similarity-based-label-set-prediction} for further details.
As an aside, binary and multi-class problems are special cases of the proposed family of similarity-based multi-label learning algorithms.
Furthermore, the binary and multi-class algorithms are recovered as special cases of $\sml$ 
when $|Y_i|=1, \, \text{ for } 1\leq i \leq \n$.
Indeed, the proposed similarity-based multi-label learning approach expresses a family of algorithms as many components are interchangeable such as the 
similarity function $\Phi$, 
normalization, 
weighting function $\Psi$ used to control the influence of the individual similarity score $S_{ij}$, 
and the sampling or sketching approach to reduce the training data.
The expressiveness and flexibility of SML enables it to be easily adapted for application-specific tasks and domains.
In addition, SML lends itself to an efficient and straightforward parallel implementation.

\subsection{Similarity-based Label Set Prediction} \label{sec:similarity-based-label-set-prediction}
We present a similarity-based approach for predicting the label set size.
For each label set $Y_i$ corresponding to a training instance $\vx_i$ in the training set $\D$, 
we set its label to $|Y_i|$, \ie, the number of labels associated with $\vx_i$.
Let $\vy = [ \, y_1 \;\, y_2 \;\, \cdots \;\, y_{\n} \, ] \in \RR^{\n}$ denote an $\n$-dimensional label vector where each $y_i = |Y_i|$ is the new transformed cardinality label of $\vx_i$ in $\D$.
The new label vector $\vy \in \RR^{\n}$ is used to predict the label set size.
In particular, the new training data is:
$\D^{\prime} = \{(\vx_i, y_i)\}, \text{ for } \, i=1,2,\ldots,\n $
\noindent
where the label set $Y_i$ of each instance is replaced by its transformed label $y_i$ that encodes the label set size $|Y_i|$ of $\vx_i$.
Furthermore, let $\Y^{\prime} = \{|Y_i|\}_{i=1}^{\n}$ denote the label space given by the transformation and $\k^{\prime} = |\Y^{\prime}|$ denote the number of unique labels (\ie, label set cardinalities).
It is straightforward to see that the above transforms the original multi-label classification problem into a general multi-class problem for predicting the label set size.

Given $\D^{\prime} = \{(\vx_1, y_1),\ldots, (\vx_\n, y_\n)\}$, the label set size of an unseen instance $\vx_i$ is predicted as follows.
First, the similarity of $\vx_i$ with respect to each training instance $(\vx_j, y_j) \in \D^{\prime}$ is derived as $\simf(\vx_i,\vx_j), \, 1 \leq j \leq \n$ and the similarities from training instances with the same set size (label) $k \in \Y^{\prime}$ are combined via addition.
More formally, the similarity of instances in $\D^{\prime}$ of the same set size (class label) $k \in \Y^{\prime}$ with respect to $\vx_i$ is:
\begin{align} \label{eq:sum-sim-from-single-class}
f_k(\vx_i) \; = \sum_{\vx_j \in \mathcal{D}^{\prime}_k} \simf\inner{\vx_i}{\vx_j}
\end{align}
\noindent where $\mathcal{D}^{\prime}_k \subseteq \D^{\prime}$ is the subset of training instances with label $k \in \Y^{\prime}$.
Therefore, we predict the set size of $\vx_i$ using the following decision function:
\begin{align}\label{eq:max-sim}
\xi(\vx_i) = &\argmax\limits_{k \in \Y^{\prime}} \;\; \sum\limits_{\vx_j \in \D^{\prime}_k} \simf\inner{\vx_i}{\vx_j} 
\end{align}
\noindent 
where $\xi(\cdot)$ is the predicted label set size for $\vx_i$.
It is straightforward to see that $\xi(\vx_i)$ is the label set size with maximum similarity.
Given the label set size $\xi(\vx_i)$, we predict the label set $\hat{Y}_i$ of $\vx_i$
by ordering the labels from largest to smallest weight based on $f_1(\vx_i), f_2(\vx_i), \ldots, f_\k(\vx_i)$ and setting $\hat{Y}_i$ to the top $\xi(\vx_i)$ labels with the largest weight.

\smallskip\noindent\textbf{Other approaches:}
Alternatively, we can infer the label set of $\vx_i$ by learning a threshold function $t : \X \rightarrow \RR$ such that:
\begin{equation} \label{eq:threshold-function}
h(\vx) = \Big\lbrace \, k \;\, | \;\, f_k(\vx) > t(\vx), \; k \in \Y \, \Big\rbrace
\end{equation}
\noindent
where $f_k(\vx)$ is the confidence of label $k \in \Y$ for the unseen test instance $\vx$.
To learn the threshold function $t(\cdot)$, we assume a linear model 
$t(\vx) = \langle \vw, f(\vx) \rangle + b$.
More formally, we solve the following problem based on the training set $\D$:
\begin{equation}\label{eq:learn-threshold-function-linear-least-squares}
\mini\limits_{\vw, b} \;\; \sum_{i=1}^{\n} \Big[ \langle \vw, f(\vx_i) \rangle + b - s(\vx_i) \Big]^2
\end{equation}
\noindent
In Eq.~\ref{eq:learn-threshold-function-linear-least-squares}, we set 
$s(\vx_i)$ as:
\begin{align}\label{eq:learn-threshold-function}
s(\vx_i) \,=\, \argmin\limits_{\tau \in \RR} \;\, 
\big|\{k \in Y_i \;\mathrm{s.t.}\; f_k(\vx_i) \leq \tau\}\big| 
\,+\, \big|\{q \in \bar{Y}_i \;\mathrm{s.t.}\; f_q(\vx_i) \geq \tau \}\big| 
\end{align}
\noindent
where $\bar{Y}_i$ is the complement of $Y_i$.
After learning the threshold function $t(\cdot)$, we use it to predict the label set $Y_i$ for the unseen instance 
$\vx_i$.
Nevertheless, any approach that predicts the label set $Y_i$ from the learned 
weights $f_1(\vx_i), \ldots, f_{\k}(\vx_i)$ can be used by $\sml$; see~\cite{tsoumakas2006multi,zhang2014review} for other possibilities.

\subsection{Complexity Analysis} \label{sec:complexity-analysis}
Given a single test instance $\vx$, the runtime of $\sml$ is $\O(\n\m \bar{\k})$ where $\n$ is the number of training instances, $\m$ is the number of attributes, and $\bar{\k} = \frac{1}{\n} \sum_{i=1}^{\n} |Y_i|$ is the average number of labels per training instance.
This is straightforward to see as $\sml$ derives the similarity between each training instance's $\m$-dimensional attribute vector.
The space complexity of $\sml$ for a single test instance $\vx$ is $\O(\k)$ where $\k$ is the number of labels.
This obviously is not taking into account the space required by $\sml$ and other methods to store the training instances and the associated label sets.
For the similarity-based set size prediction approach, the time complexity is only $\O(\n\m)$ since the label set size with maximum similarity can be maintained in $o(1)$ time. 
However, the approach uses $\O(\k^{\prime})$ space where $\k^{\prime} \leq \k$.

It is straightforward to incorporate a sampling mechanism into the approach to further improve the time and space requirements. 
In particular, given a new test instance $\vx$ we can sample a small fraction of training instances denoted by $\D_s$ via an arbitrary distribution $\mathbb{F}$ and use this smaller set for predicting labels for $\vx$.

%======================================
% EXPERIMENTS
%======================================
\section{Experiments} \label{sec:exp}
This section investigates the practical significance of $\sml$ for multi-label classification.
In particular, we evaluate $\sml$ against a wide variety of multi-label algorithms including:
\begin{compactitem}
\item \textsc{Ml-knn}~\cite{ml-knn}: A kNN-based multi-label approach that uses Euclidean distance to find the top-k instances that are closest. \textsc{Ml-knn} was shown to perform well for a variety of multi-label problems.
\item \textsc{BoosTexter}~\cite{boostexter}: A boosting-based multi-label algorithm called \textsc{BoosTexter}.
\item \textsc{Adtboost.mh}~\cite{adtboost}: A multi-label decision tree approach.
\item \textsc{Rank-svm}~\cite{ranksvm}: 
A multi-label SVM approach based on ranking.
\end{compactitem}
For \textsc{BoosTexter} and \textsc{Adtboost.mh} we use 500 and 50 boosting rounds respectively since performance did not change with more rounds (which is consistent with~\cite{ml-knn}).
For \textsc{Rank-svm} we use polynomial kernels with degree $8$ which performs best as shown in~\cite{ranksvm,ml-knn}.
Unless otherwise mentioned, our approach uses the RBF similarity function in Eq.~\eqref{eq:rbf}; the RBF hyperparameter is learned automatically via k-fold cross-validation on $10\%$ of the labeled data. % validation set
In this work, we systematically compare the multi-label learning algorithms using data from different domains.

\subsection{Gene functional classification} \label{sec:exp-gene-function-classification}
The first multi-label learning task we evaluate is based on predicting the functions of genes from Yeast Saccharomyces cerevisiae - a widely studied organism in bioinformatics~\cite{pavlidis2000combining}.
Each gene may take on multiple functional classes.
In this investigation, we used the Yeast data from~\cite{ranksvm,pavlidis2000combining}.
Each gene consists of a concatenation of micro-array expression data and phylogenetic profile data.
Following Elisseeff~\etal~\cite{ranksvm}, we preprocess the data such that only the known structure of the functional classes are used.
The resulting multi-label yeast data consists of $\n=2417$ genes 
where each gene is represented by a $103$-dimensional feature vector. 
There are $\k=14$ labels denoting the functional classes.

\begin{table*}[h!]
\centering
\setlength{\tabcolsep}{5.5pt}
\ra{1.25}
\scriptsize
\caption{Experimental results for each multi-label learning algorithm on the yeast data (mean$\pm$std).}
\vspace{1mm}
\label{table:yeast-results}
\small
\scriptsize
\fontsize{8.0}{9.0}\selectfont
\begin{tabularx}{1.0\linewidth}{r l llll H}
\toprule
\TTT\BBB
Evaluation criterion 
& $\sml$ & \textsc{Ml-knn}~\cite{ml-knn} & \textsc{BoosTexter}~\cite{boostexter} & \textsc{Adtboost.mh}~\cite{adtboost} & \textsc{Rank-svm}~\cite{ranksvm} &
\\ \midrule

Hamming loss ($\downarrow$)  &  \textbf{0.193 $\pm$ 0.013}  &
0.194 $\pm$ 0.010  &  
0.220 $\pm$ 0.011  &  
0.207 $\pm$ 0.010  &  
0.207 $\pm$ 0.013  &  
\\

One-error ($\downarrow$) &  \textbf{0.220 $\pm$ 0.021}  &  
0.230 $\pm$ 0.030  &  
0.278 $\pm$ 0.034  &  
0.244 $\pm$ 0.035  &  
0.243 $\pm$ 0.039  &  
\\

Coverage ($\downarrow$) &  \textbf{6.082 $\pm$ 0.184}  &  
6.275 $\pm$ 0.240  &  
6.550 $\pm$ 0.243  &  
6.390 $\pm$ 0.203  &  
7.090 $\pm$ 0.503  &  
\\

Ranking loss ($\downarrow$) &  
\textbf{0.155 $\pm$ 0.011}  &
0.167 $\pm$ 0.016  &  
0.186 $\pm$ 0.015  &  
N/A &
0.195 $\pm$ 0.021  &  
\\

Average precision ($\uparrow$) &  
\textbf{0.783 $\pm$ 0.016}  & 
0.765 $\pm$ 0.021  &  
0.737 $\pm$ 0.022  &  
0.744 $\pm$ 0.025  &  
0.749 $\pm$ 0.026  &  
\\

\bottomrule
\end{tabularx}
\end{table*}

We use 10-fold cross-validation and show the mean and standard deviation.
Experimental results for $\sml$ and other multi-label learning algorithms are reported in Table~\ref{table:yeast-results}.
Notably, all multi-label algorithms are compared across a wide range of evaluation metrics.
The best result for each evaluation criterion is shown in bold.
In all cases, our approach outperforms all other multi-label learning algorithms across all 5 evaluation criterion.
Furthermore, the variance of $\sml$ is also smaller than the variance of other multi-label learning algorithms in most cases.
This holds across all multi-label learning algorithms for coverage, average precision, and ranking loss.\footnote{Note the \textsc{Adtboost.mh}~\cite{adtboost} program does not provide ranking loss.}

\subsection{Scene image classification} \label{sec:exp-scene-classification}
The second multi-label learning task we evaluate $\sml$ for is natural scene classification using image data.
In scene classification each image may be assigned multiple labels representing different natural scenes such as an image labeled as a mountain and sunset scene.
Therefore, given an unseen image the task is to predict the set of scenes (labels) present in it.
The scene data consists of 2000 images where each image contains a set of manually assigned labels.
There are $\k=5$ labels, namely, desert, mountains, sea, sunset, and trees.
Each image is represented by a $294$-dimensional feature vector derived using the approach in~\cite{boutell2004learning}.

\begin{table*}[h!]
\centering
\setlength{\tabcolsep}{5.5pt}
\ra{1.25}
\scriptsize
\caption{Results of the multi-label learning algorithms for natural scene classification (mean$\pm$std).}
\vspace{1mm}
\label{table:natural-scene-results}
\small
\scriptsize
\fontsize{8.0}{9.0}\selectfont
\begin{tabularx}{1.0\linewidth}{r l llll H}
\toprule
\TTT\BBB
Evaluation criterion 
& $\sml$ & \textsc{Ml-knn}~\cite{ml-knn} & \textsc{BoosTexter}~\cite{boostexter} & \textsc{Adtboost.mh}~\cite{adtboost} & \textsc{Rank-svm}~\cite{ranksvm} &
\\ \midrule

Hamming loss ($\downarrow$) &  \textbf{0.140 $\pm$ 0.009}  & 
0.169 $\pm$ 0.016  &  
0.179 $\pm$ 0.015  &  
0.193 $\pm$ 0.014   &  
0.253 $\pm$ 0.055   &  
\\

One-error ($\downarrow$) &  \textbf{0.252 $\pm$ 0.026}  &  
0.300 $\pm$ 0.046  &  
0.311 $\pm$ 0.041  &  
0.375 $\pm$ 0.049  &  
0.491 $\pm$ 0.135 &
\\

Coverage ($\downarrow$) &  0.984 $\pm$ 0.112  &  
\textbf{0.939 $\pm$ 0.100}  &  
\textbf{0.939 $\pm$ 0.092}  &  
1.102 $\pm$ 0.111  &  
1.382 $\pm$ 0.381 &
\\

Ranking loss ($\downarrow$) &  \textbf{0.164 $\pm$ 0.020}  &  
0.168 $\pm$ 0.024  &  
0.168 $\pm$ 0.020  &  
N/A  &  
0.278 $\pm$ 0.096 &
\\

Average precision ($\uparrow$) &  \textbf{0.852 $\pm$ 0.016}  &  
0.803 $\pm$ 0.027  &  
0.798 $\pm$ 0.024  &  
0.755 $\pm$ 0.027  &  
0.682 $\pm$ 0.093
\\

\bottomrule
\end{tabularx}
\end{table*}

We use 10-fold cross-validation and show the mean and standard deviation.
The experimental results of $\sml$ and the other multi-label algorithms using the natural scene classification data are reported in Table~\ref{table:natural-scene-results}.
The best result for each evaluation criterion is in bold.
From Table~\ref{table:natural-scene-results}, it is obvious that $\sml$ outperforms all other multi-label algorithms on all but one evaluation criterion, namely, coverage.
In terms of coverage \textsc{Ml-knn} and \textsc{BoosTexter} are tied and have slightly lower coverage than $\sml$.

\section{Conclusion} \label{sec:conc}
We have described a general framework for similarity-based multi-label learning called $\sml$ that gives rise to a novel class of methods for the multi-label problem.
Furthermore, we also presented a similarity-based approach for predicting the label set size. 
Experiments on a number of data sets 
demonstrate the effectiveness of $\sml$ as it compares favorably to existing methods across a wide range of evaluation criterion.

\setlength{\bibsep}{4pt plus 0.3ex}
\bibliography{paper-arxiv}
\bibliographystyle{abbrvnat}

\end{document}

%% file: preamble.tex
\usepackage[notheorems]{rr-math}
\usepackage{nicefrac}

\usepackage{amsmath,lipsum}
\renewcommand{\pm}{\mathbin{\smash{%
\raisebox{0.35ex}{%
            $\underset{\raisebox{0.5ex}{$\smash -$}}{\smash+}$%
            }%
        }%
    }%
}

\usepackage[table]{xcolor}

\usepackage{xcolor}
\usepackage{verbatim}
\usepackage{wrapfig}
\usepackage{booktabs}

\usepackage{graphicx}
\usepackage{multirow}
\usepackage{rotate}
\usepackage{subfigure}
\usepackage{epstopdf}
\usepackage{paralist}
\usepackage{url}
\usepackage{graphicx}
\usepackage{wrapfig}
\usepackage{graphicx}
\usepackage{amssymb}
\usepackage{amsmath}
\usepackage{amsfonts}

\newcommand{\todo}[1]{}

\newcommand{\incomplete}[1]{}

\usepackage{tabularx}
\usepackage{booktabs}
\usepackage{algorithm}
\usepackage{algorithmicx}
\usepackage{algpseudocode}

\algblockdefx[parallel]{ParFor}{EndPar}[1][]{$\textbf{parallel for}$ #1 $\textbf{do}$}{$\textbf{end parallel}$}
\algrenewcommand{\alglinenumber}[1]{\fontsize{6.5}{7}\selectfont#1}
\algblockdefx[parallel]{parfor}{endpar}[1][]{$\textbf{parallel for}$ #1 $\textbf{do}$}{$\textbf{end parallel}$}
\algrenewcommand{\alglinenumber}[1]{\scriptsize#1:}

\usepackage{nicefrac}

\algnotext{EndIf}
\algnotext{EndProcedure}
\ifpdf
  \usepackage{epstopdf}
\fi
\graphicspath{{./}{./figures/}}

\graphicspath{{graphics/}{../graphics/}{./}}

\usepackage{array}
\newcolumntype{P}[1]{>{\centering\arraybackslash}p{#1}}
\newcolumntype{M}[1]{>{\centering\arraybackslash}m{#1}}

\relax

\let\hat\widehat

\usepackage{wrapfig}
\usepackage{xspace}
\newcommand{\etal}{\emph{et al.}\xspace}

\newlength{\commentWidth}
\newcommand{\bspacing}{\begin{spacing}{1.4}}
\newcommand{\espacing}{\end{spacing}}

\definecolor{plotblue}{RGB}	{30,144,255}
\definecolor{plotgreen}{RGB}	{50,205,50}
\definecolor{plotred}{RGB}	{220,20,60}

\definecolor{myyellow}{RGB}{255,255,204}
\definecolor{myred}{RGB}{255,204,204}
\definecolor{myblue}{RGB}{204, 255, 255}
\definecolor{mygreen}{RGB}{204, 255, 204}

\definecolor{gray}{RGB}{150,150,150}
\definecolor{theblue}{RGB}{0,0,180}
\newcommand*\hrulefillvar[1][0.4pt]{\leavevmode\leaders\hrule height#1\hfill\kern0pt}

\newcommand{\be}{\begin{equation}}
\newcommand{\ee}{\end{equation}}
\newcommand{\bea}{\begin{eqnarray}}
\newcommand{\eea}{\end{eqnarray}}
\newcommand{\bit}{\begin{itemize}}
\newcommand{\eit}{\end{itemize}}

\definecolor{lightgray}{rgb}{0.93,0.93,0.93}

\definecolor{lightblue}{rgb}{0.5,0.90,1.0}
\definecolor{lightgreen}{rgb}{0.5,0.92,0.5}
\definecolor{lightred}{rgb}{0.98,0.5,0.5}
\definecolor{lightyellow}{rgb}{1,0.90,0.40}

\newcommand\TTT{\rule{0pt}{3.2ex}}
\newcommand\BBB{\rule[-1.4ex]{0pt}{0pt}}

\usepackage{booktabs}
\usepackage{comment}

\newcommand{\eat}[1]{}

\definecolor{myyellow}{RGB}{255,255,204}
\definecolor{myred}{RGB}{255,204,204}
\definecolor{myblue}{RGB}{0,200,255}
\definecolor{mygreen}{RGB}{80,220,80}

\newcommand{\eg}{\emph{e.g.}}
\newcommand{\ie}{\emph{i.e.}}

\newcommand{\inner}[2]{\left\langle #1,#2 \right\rangle}

\algblockdefx[parallel]{parallelfor}{parallelend}
[1][]{\textbf{parallel for} #1}
{\textbf{end parallel}}

\RequirePackage{mathtools}
\RequirePackage{color}
\RequirePackage{listings}

\newcommand{\ds}\displaystyle
\newcommand{\mbb}\mathbb
\newcommand{\mc}\mathcal
\newcommand{\del}\nabla

\newcommand{\beqstar}{\begin{eqnarray*}}
\newcommand{\eeqstar}{\end{eqnarray*}}

\definecolor{thegreen}{rgb}{0,.5,0}
\definecolor{idea}{rgb}{0,.6,0.1}
\definecolor{problem}{rgb}{0.7,0,0.1}
\definecolor{comment-green}{rgb}{0,.3,0}
\definecolor{theblue}{rgb}{0,0,.8}
\definecolor{light-gray}{gray}{0.98}
\definecolor{comment-color}{rgb}{0,0,.8}
\definecolor{string-color}{rgb}{0,.75,0}
\definecolor{border-blue}{rgb}{0,0,.6}

\usepackage{epsfig}

\newcommand{\ra}[1]{\renewcommand{\arraystretch}{#1}}

\usepackage{ragged2e}

\usepackage{setspace}
\graphicspath{{./}{./images/}}
\newcolumntype{H}{>{\setbox0=\hbox\bgroup}c<{\egroup}@{}}

\definecolor{orange}{rgb}{1,0.5,0}
\definecolor{gray}{RGB}{20,20,20}
\definecolor{greencm}{RGB}{0,153,0}

\usepackage[T1]{fontenc}

\definecolor{gray}{RGB}{150,150,150}
\definecolor{theblue}{RGB}{0, 20, 159} 
\definecolor{thedarkblue}{RGB}{0,0,120} 
\usepackage{hyperref}

  \definecolor{mydarkblue}{rgb}{0,0.08,0.45} 
  \hypersetup{%
    colorlinks=true,
    linkcolor=mydarkblue,
    citecolor=mydarkblue,
    filecolor=mydarkblue,
    urlcolor=mydarkblue,
    pdfview=FitH}

\usepackage{ifpdf}
\usepackage{paralist}
\usepackage{tabularx}
\usepackage{booktabs}
\usepackage{multirow}
\usepackage{graphicx}
\usepackage{amssymb}
\usepackage{amsmath}
\usepackage{mathdots}
\usepackage{lscape}

\usepackage{subfigure}
\usepackage{array}
\usepackage{url}

%% file: preamble-notation-new.tex
\newcommand{\sml}{\textsc{Sml}}

\providecommand{\RR}{\mathbb{R}} %
\providecommand{\Std}{\mathbb{S}} %

\providecommand{\Y}{\ensuremath{\mathcal{Y}}}
\providecommand{\X}{\ensuremath{\mathcal{X}}}
\providecommand{\D}{\ensuremath{\mathcal{D}}}

\providecommand{\m}{\ensuremath{M}} % 
\providecommand{\n}{\ensuremath{N}}  % 
\renewcommand{\k}{\ensuremath{K}}  % 

\providecommand*{\mini}{\mathop{\mathrm{minimize}}}

\renewcommand{\argmax}{\operatornamewithlimits{\arg \; \max}}
\renewcommand{\argmin}{\operatornamewithlimits{\arg \; \min}}

\renewcommand{\phi}{\ensuremath{\Phi}}

\newcommand{\simf}{\ensuremath{\Phi}}

\providecommand{\E}{\ensuremath{E}}

\providecommand{\X}{\ensuremath{X}}  %

\providecommand{\p}{\ensuremath{p}} % 

\providecommand{\p}{\ensuremath{p}}

\renewcommand{\O}{\ensuremath{\mathcal{O}}}
\renewcommand{\bigO}[1]{\ensuremath{\mathcal{O}(#1)}}